# Machine learning techniques for the Schizophrenia diagnosis: A comprehensive review and future research directions

Shradha Verma[1], Tripti Goel[1], M Tanveer[2], Weiping Ding[3], Rahul Sharma[1] and R Murugan[1]

[1*]Biomedical Imaging Lab, Department of Electronics and Communication Engineering, NIT Silchar, Assam, 788010, India.
[2]Department of Mathematics, Indian Institute of Technology Indore, Simrol, Indore, MP, 453552, India.
[3]School of Information Science and Technology, Nantong University, Nantong, 226019, China.

Contributing authors: shradha_rs@ece.nits.ac.in; triptigoel@ece.nits.ac.in; mtanveer@iiti.ac.in; dwp9988@163.com; rahul_rs@ece.nits.ac.in; murugan.rmn@ece.nits.ac.in;

**Abstract**

Schizophrenia (SCZ) is a brain disorder where different people experience different symptoms, such as hallucination, delusion, flat-talk, disorganized thinking, etc. In the long term, this can cause severe effects and diminish life expectancy by more than ten years. Therefore, early and accurate diagnosis of SCZ is prevalent, and modalities like structural magnetic resonance imaging (sMRI), functional MRI (fMRI), diffusion tensor imaging (DTI), and electroencephalogram (EEG) assist in witnessing the brain abnormalities of the patients. Moreover, for accurate diagnosis of SCZ, researchers have used machine learning (ML) algorithms for the past decade to distinguish the brain patterns of healthy and SCZ brains using MRI and fMRI images. This paper seeks to acquaint SCZ researchers with ML and to discuss its recent applications to the field of SCZ study. This paper comprehensively reviews state-of-the-art techniques such as ML classifiers, artificial neural network (ANN), deep learning (DL) models, methodological fundamentals, and applications with previous studies. The motivation of this paper is to benefit from finding the research gaps that may lead to the development of a new model for accurate SCZ diagnosis. The paper concludes with the research finding, followed by the future scope that directly contributes to new research directions.

## 1 Introduction

Schizophrenia (SCZ) is a serious mental condition distinguished by delusions and hallucinations associated with a lack of contact with reality, flat emotions, anhedonia, loss of desire (avolition), terrible speech (alogia), social disengagement, and cognitive impairment, all of which have a significant impact on the individual and community. Around 20 million people worldwide are affected by this disease [1]. The specific cause of SCZ is still unknown, but some factors like stressful life events, drugs, etc., and their combinations are assumed to cause SCZ. Broadly, its symptoms are categorized into positive and negative [2]. Positive symptoms include hallucinations, delusions, disorganized thinking, etc., and negative symptoms are the ones that impact a person's workability, lack of motivation, inability to concentrate, reduced range of emotions, etc. Early diagnosis of the SCZ helps slow down the disease's progression by treating it at the initial stage.

Neuroimaging plays a significant role in showing both functional and structural changes in the human brain [2]. Brain anatomy using structural magnetic resonance imaging (MRI) technique provides a reliable assessment of SCZ diagnosis. The non-invasive approach involves strong magnetic fields, magnetic field gradients, and radio waves to generate brain images. On the other hand, functional magnetic resonance imaging (fMRI) estimates brain activity by detecting transitions of blood flow. The technique relies on blood oxygenation level-dependent (BOLD) imaging. Diffusion tensor imaging (DTI) is a potential tool for assessing microstructural changes or variations with neuropathology and therapy. The diffusion tensor may describe the amplitude, the extent of anisotropy, and the directional diffusion. Another popular modality, electroencephalogram (EEG), acquires the brain signals based on the brain's frequencies (delta, alpha, beta, and gamma frequencies) that aid in analyzing cognitive behavior.

High-precision analysis of voluminous imaging data requires the use of computational approaches. Once the most important data has been retrieved from the scans, a machine learning (ML) based classification approach may be used to calculate the disease's likelihood. Support vector machines (SVM), Decision Tree (DT), and Random Forest (RF) are all examples of the approach to ML and are widely adopted. In recent years machine learning (ML) algorithms have played an important role in disease diagnoses like tumors, cancer, neurodegenerative disorders, and psychiatric disorders [3]. Exploring brain functionality and making precise predictions are made possible by applying ML methods to neuroimaging data. With its roots in Artificial intelligence, machine learning creates algorithms that enable computers to "learn" new skills by analyzing existing ones.

Learning has the meaning of discovering patterns in data using statistical analysis. Machine learning's strength in deducing meaningful patterns from large amounts of data is put to good use in the context of pattern recognition challenges. However, handcrafted features must be fed to the ML model for improved learning. Deep learning (DL), a subset of ML, has been an acknowledged approach due to its manual feature extraction and selection capabilities. DL has been widely adopted for disease classification, segmentation, data synthesis, and retrieval [4].

The novelty of this study aims to assess the evidence for the use of ML approaches in SCZ diagnostic discrimination during the past three decades. No systematic reviews exploring these characteristics have been found to our knowledge. This study provides an argumentative comparison with state-of-the-art techniques, highlighting their advantages, limitations, and comparisons. The motivation of this paper is to help the researchers understand the fundamentals of ML algorithms and the different neuroimaging modalities that assist in understanding plane views on brain structure, functioning, and interconnectivity.

Following are the contributions of the review paper:

1. Brief introduction and clinical findings of the prevalent neuroimaging modalities for SCZ diagnosis.
2. Discussion on the available datasets accessed for SCZ diagnosis.
3. Introduction of ML, ANN, & DL algorithms for diagnosing SCZ.
4. Discussion on the advantage and limitations of various algorithms for SCZ detection.
5. Prospects for the future application of ML/DL in diagnosing SCZ, which other researchers can implement.

The paper is organized as follows, Section 2 addresses the search strategy, followed by section 3, which discusses preliminaries. Section 4 introduces the application of ML techniques for SCZ diagnosis. In contrast, section 5 discusses challenges and future scope, followed by a conclusion in section 6.

## 2 Search Strategy

A total of 657 Papers are searched from PubMed and Google Scholar using the keywords SCZ and neuroimaging. In the first screening, papers consisting of small datasets that did not match the current modality, feature extraction, and ML or DL techniques are excluded. After initial screening, twenty-nine papers based on DL and twenty-eight ML papers were selected. The flow chart of the search strategy and year-wise statics of the paper bar graph is shown in Figure 1.

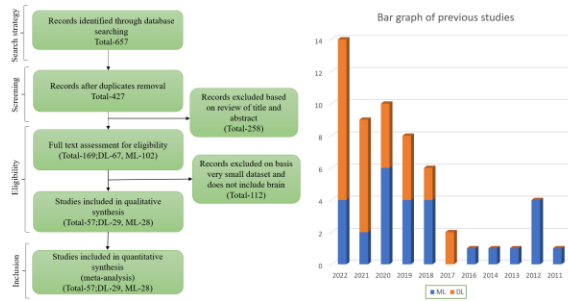

**Fig. 1** Flowchart of search strategy and statistics of papers on schizophrenia

## 3 Preliminaries

### 3.1 Pre-Processing

MR imaging is advantageous because of its superior soft-tissue contrast and lack of ionizing radiation. However, there is no standard scale for assessing image intensity, so these scans are usually more challenging to evaluate. Additionally, artifacts from various sources must be considered before implementing a computer-aided diagnosis approach. The primary MRI artifacts involve the scanner's variation, motion, and in-homogeneity. Based on previous studies, popular preprocessing steps are:

1. Image Realignment: It helps to remove the various artifacts from MRI.
2. Image Normalization: It normalizes the image intensities for all voxels.
3. Image Registration: It aligns all the scans geometrically. Slice level matching is executed to attain exact dimensions for scans. The standard MNI-152 template is widely used for image registration.

Voxel-based Morphometric tools such as Statistical Parametric Mapping 12 (SPM 12) [5], FMRIB Software Library (FSL) [6], Analysis of Functional NeuroImages (AFNI) [7], etc., have been used to preprocess the MRI volume data. After preprocessing, slice extraction is performed by extracting 2$D$ slices from 3$D$ neuroscans to reduce the computational complexity.

### 3.2 Principle of neuroimaging techniques

Human bodies contain 60-70% water, and hydrogen atoms are magnetically sensitive. Dipole rotation creates a magnetic field equal to the magnetic moment within the particle. Without an external magnetic field, protons are grouped randomly, nullifying the magnetic field. When the patient's body approached the main magnetic field ($B_o$), the hydrogen atom protons aligned in the longitudinal direction, most of them parallel to the anti-parallel field. Thus, the net magnetization vector is in the $B_o$ field or z-axis, called precession, and proton moments per second are associated with Larmor frequency. Anti-parallel protons absorb energy and emit electromagnetic (EM) energy while returning to their original location when Larmor frequency radio waves are delivered. MRI uses EM energy to create images.

The brain has a neurotransmitter chemical that conveys signals in the brain's distinct parts and is accountable for brain connectivity. In SCZ, an imbalance of the brain chemicals or neurotransmitters, such as dopamine, glutamate, serotonin, etc., has been seen. The imbalance of these chemicals affects brain connectivity. Due to this disturbance in the brain, white matter (WM), interconnection of dendrites, and synapses are also altered. Ventricle enlargement increases over time. The brain's language processing (superior temporal gyrus and its connections) region is mainly affected by the reduction in the hippocampus (HP) and thalamic (Th) volumes increasing the globus pallidus (Gd) volume found in the schizophrenic brain as shown in Figure 2. On the other hand, fMRI measures map brain activity. Also, this neuroimaging differentiates between oxygenated and deoxygenated blood flow. The hydrogen atom emits an EM signal in magnetic field presence. MR signal, in an indirect way, is

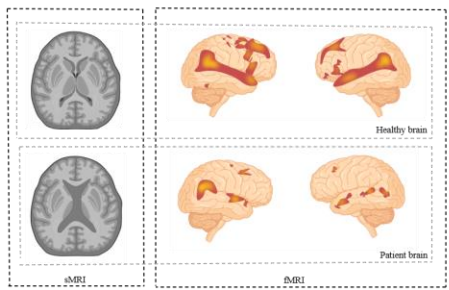

**Fig. 2** sMRI and fMRI modalities to show the difference between SCZ patients and healthy control (figure are created with BioRender.com)

sensitive to the neural activity and magnetic properties of oxygen. Oxygen binds with hemoglobin, responsible for blood flow. The level of oxygenated blood flows in the brain is proportional to the change in neural activity in that part. More oxygen levels mean stronger signals as shown in Figure 2. DTI principle includes the random thermal motion of water molecules." In other words, it fundamentally uses the diffusion of water as a probe to evaluate the anatomy of a brain network. Due to the heterogeneity of the tissue, water molecules do not diffuse uniformly in all directions (anisotropic diffusion). Because there are proportionally fewer impediments along the fiber, water molecules should move more quickly along the axon than they would if they were moving vertically to the fiber. Hence, anisotropic diffusion can provide an entirely different image contrast based on the axonal direction, which is particularly helpful in identifying significant brain structures.

Similarly, EEG looks for an irregular electrical activity or brain waves in your brain. During the process, the scalp will be covered in electrodes, which are tiny metal discs with flimsy wires. The brain's activity generates minute electrical charges that are detected by the electrodes. When recording voltage differences using a pair of electrodes for an EEG, the principle of differential amplification is used to compare one active exploration electrode location with a different nearby or distant reference electrode. Moreover, a computer screen graph or a recording that can be printed out on paper shows the amplified charges. Your healthcare provider then interprets the reading.

## 3.3 Introduction of Machine Learning techniques

This section introduces the traditional ML, ANN, and trending DL techniques for SCZ diagnosis.

### 3.3.1 Machine Learning

ML is vital for every researcher who wants to turn raw data into patterns and forecasts. Several ML algorithms has been available and widely adopted by the community. Vapnik et al. [8] introduced a support vector machine (SVM), a supervised machine learning method designed for classification and regression problems. Broadly, SVM selects the hyperplane which separates the dataset with maximum distance from the decision boundary. The higher the margin better will be the result obtained. The geometric representation of SVM is shown in Figure 3(A). Thus, multi-dimensional datasets fit with maximal margin into different classes. If the dataset is linearly separable, a linear-type hyperplane is used. However, if the dataset is non-linear or linearly inseparable, different kernel types were introduced to classify the dataset perfectly. The classifier's performance greatly depends on properly selecting the kernel type and its parameters [9]. For higher dimension hyperplane, the decision function is given as $f(x) = w^T X + b$, where $w$ is the vector orthogonal to the hyperplane, $b$ is the constant or bias, and $X$ are the variables.

Random forest (RF) [10] uses an ensemble of decision trees (DT) to improve the performance of DT. Bagging or bootstrap aggregation are used to train the algorithm formed by the RF. Bagging is a meta-algorithm that increases the ACC of ML methods by grouping them together. It determines the outcome based on DT predictions as shown in Figure 3(B).

Artificial neural network (ANN) [11] has input, output, and hidden layers to simulate human brain. Input neurons or cells are linked to hidden layers by a channel. Each channel assigns a weight whose product with input value is sent to the next layer (hidden layer). Next, output relies on node weight or likelihood. Forwarding propagation forwards input via a hidden layer. The ANN constantly learns its training data, but after forwarding propagation, the real value and anticipated value are computed, and the difference is an error. Error size influences direction. Backpropagation (BP) enables the device to learn on training data, as seen in Figure 3(C).

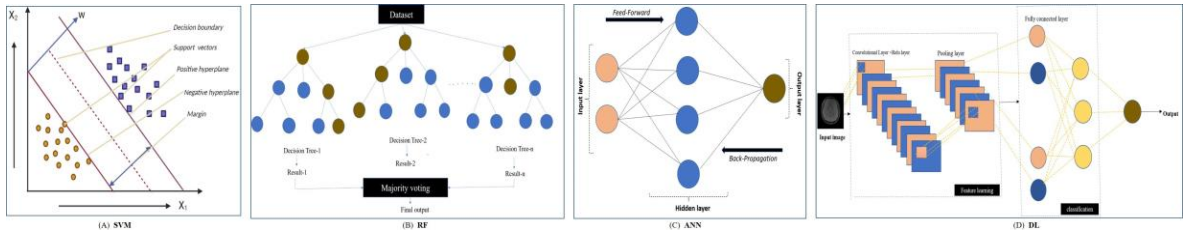

**Fig. 3** Block representation of (A) SVM (B) RF (C) ANN and (D) DL approaches.

### 3.3.2 Deep Learning

DL approach is a subset of ML [12]. ML techniques require manual feature extraction, whereas the DL technique extracts feature automatically as it consists of multiple hidden layers. DL algorithms are data-hungry and require a huge amount of data for classification problems. The basic architecture of the general DL modal is shown in Figure 3(D). Input, convolutional, max pooling, full-connected, and classification layers are CNN building components. All picture pixels multiply with the convolution kernel or filter. In this layer, shifting and multiplication create the final complicated feature map. ReLU layers cancel out negative pixels. The convolution layer extracts high-level input image features such as color, edges, etc. The pooling layer down-samples certain data to keep it. Max pooling collects the maximum pixel value from filters, whereas average pooling averages the feature map. After feature mapping, a fully connected layer transforms the feature matrix into a vector, then a softmax or sigmoid activation function classifies the input [13].

## 4 Application of Machine Learning for SCZ diagnosis

The section includes the application and contribution of the previous works on ML, ANN, and DL.

### 4.1 Dataset

Several publicly available neuroimaging data for analysis of SCZ patients have been included. One of the main online datasets is SchizConnect [14].

A total of 1,298 neuro scans from subjects are available in this database. Inside SchizConnect, **5** multicenter datasets are separately available, and their demographic details are given in Table 1. The above data sets have collected to investigate the brain metabolism of SCZ patients (BrainGluSchi) and contained both structural and functional imaging Centre of biomedical research excellence (COBRE) and Mind clinical imaging consortium (MCIC)). All the datasets are available over variable strength fields of 1.5T and 3T. COBRE investigates the brain processes of SCZ by combining neuroimaging techniques with clinical, cognitive, and genetic testing. BrainGluSchi is prospective longitudinal research looking at glutamate plus glutamine and other standard neuro-metabolites in a wide group of SCZ patients and healthy control (HC) participants. MCIC data provides extensive clinical, anatomical, functional, and diffusion-weighted neuro scans. Northwestern University Schizophrenia Data and Software Tool (NUSDAST) [15] consist of neuroimaging scans acquired from subjects like SCZ and HC with a two-year follow-up. Similarly, Stanford Center for Reproducible Neuroscience established the Open-Neuro [16] platform to promote and enhance data sharing and has included analysis of raw MRI data.

**Table 1** Demographics details about the four publicly available datasets accessed for SCZ diagnosis

| Dataset      | No. of scans | Average Age | Age Range |
| ------------ | ------------ | ----------- | --------- |
| COBRE        | 184          | 38          | 18-66     |
| MCIShare     | 204          | 34          | 18-61     |
| BrainGluSchi | 175          | 37          | 16-65     |
| NUSDAST      | 451          | 33          | 14-66     |
| OpenNeuro    | 284          | -           | -         |

## 4.2 Machine Learning

This section addresses the contribution of previous papers whose results appear promising for many clinicians and aid in SCZ diagnoses, as shown in Table 2. The following discussion provides insights into the previous work done for the two-dimensional approaches. Filipovych et al. [17] have studied aging, pathological changes, and clustering subproblems of patients. The authors observed the multivariate patterns using the Joint maximum-margin classification and clustering (JMMCC) classifier, achieving a confidence interval (CI) of 95%. Ulas et al. [18] have employed multiple kernel learning (MKL) to elect SCZ in different ROI regions like Amygdala (AD), TL, and Th. The author advised Entorhinal Cortex region is directly relevant to SCZ. However, ROI-based slices has been extracted, which leads to a reduced feature scale. ArivuSelvan and Moorthy [19] have considered Th and its subnuclei as the ROI. The author has done preprocessing steps that improve image quality and achieved a mean square error (MSE) of 0.83. However, results may exhibit variation when different ROI region is considered. Park et al. [20] have used random features of 642 bilateral hippocampal subfields. Authors have opted for several automatic segmentation techniques for the classification, followed by validation. Results achieved an ACC of 82.1%. Febles et al. [21] have tried to identify schizophrenia cases using an MKL-SVM approach with an ACC of 86%. Rosa et al. [22] has opted the RF classifier. They extracted HP and DLPFC features for twenty SCZ and HC cases whose results achieved an AUC of 0.85.

Ulas et al. [23] have introduced the three-step approach. Initially, they have taken multiplication factors for classification to improve accuracy (ACC). Secondly, it reduced the subsets of the region of interest (ROIs). Lastly, correlation analysis has been done to measure the overlap information. This yielded an optimum ACC of 83.3% . However, its ACC depends on the selection of the ROIs subset. Castellani et al. [24] have taken non-linear SVM for automatic classification. The dorsolateral prefrontal cortex (DLPFC) is a structural marker for detecting SCZ cases. Modinos et al. [25] have presented changes in the brain network within the emotional circuitry. Arbabshirani et al. [26] have utilized resting-state functional network connectivity features to distinguish SCZ subjects from HC. Results have exhibited an ACC of 88%; however, the best ACC is obtained using KNN. Castro et al. [27] have analyzed the complex fMRI data using MKL. Besides, work can extend to variable MKL with a loss function.

Jafri and Calhoun [28] have employed a feed-forward neural network to detect SCZ and also examined SCZ sub-types entirely following diagnostic and statistical manual (DSM) fourth measures in FL, parietal lobe (PL), OL, and TL brain's regions. Bae et al. [29] have observed the local and global connectivity of SCZ patients and found that activity connection is different in SCZ patients and HC. Using SVM and 10-fold cross-validation result reached a prediction ACC of 91.2%. Chin et al. [30] work has used spatial regularized SVM to classify SCZ subjects. In another study, Nimkar and Kubak [31] have compared the classification result of different algorithms such as SVM, K-Nearest Neighbor (KNN), RF, and LDA. The paper has used feature extraction and dimensionality reduction methodology to increase the efficiency and ACC of different ML techniques. Pierrefeu et al. [32] have employed a method to automatically identify the fMRI period of patients who suffer from hallucinations at the resting state. Using transformer-based Visual Exploration Network (TV-ENet) and SVM, the complex process of hallucination has been defined in his work.

Similarly, Li et al. [33, 34] have done a classification Linear Discriminant Analysis (LDA) of SCZ and HC achieved 76.34% ACC. Liu et al. [35] have proposed a modified hierarchical brain network to classify the SCZ individuals from the HC. Sutcubasi et al. [36] have investigated anatomical brain connectivity for 144 SCZ and 154 HC using spatial statics and ML techniques. The result has achieved a performance of 79.3%. Vyškovský et al. [37] have extracted the whole-brain features field on voxel-based morphometry (VBM) and deformation-based morphometry (DBM) schemes using Multilayer perceptron (MLP) classifier. The feature extraction methods help to achieve maximum ACC. Chen et al. [38] have discussed an ML framework that included two-sample t-tests to understand the variation in groups, firstly, coarse to fine features selected with recursive feature elimination (RFE). Afterward, the SVM model selected the grey matter (GM) and WM and provided an ACC of 85%. Lei et al. [39] have taken both MRI and fMRI data from 296 SCZ and 452 HC subjects. The author demonstrated that functional connectivity yielded more promising results than structural data, and coupling these data presents the highest ACC.

In contrast, Vieira et al. [40] have segmented MRI scan into surface-based regional volumes, gray matter volume (GMV), and cortical thickness. Dimensions of the segmented volumes are reduced using principal component analysis (PCA), and the extracted features are classified using four ML classifiers. The proposed algorithm requires many preprocessing steps for image segmentation, feature extraction, and classification and hence needs more computational power and storage memory. Yang et al. [41] have shown inhibition deficits in SCZ patients. Moreover, they analyzed stop-signal tasks using SVM, which helps to classify the SCZ patients and HC. Yassin et al. [42] have also used different classifiers like SVM, Adaboost (AB), and others to classify the subjects of SCZ. The result has demonstrated that SVM and Logistic regression (LR) yield the most favorable result. Lin et al. [43] have classified the 99 SCZ patients using multi-kernel SVM, and the result has shown 94.86% ACC. Using fMRI modality, Algumaei et al. [44] have employed resting-state fMRI data for schizophrenia detection. Features like HP, temporal lobe (TL), and frontal lobe (FL) were extracted using the SVM with an ACC of 98.5%. Tanveer et al. [45] have evaluated different ML modes to provide the understanding about the use of various classification algorithms, as well as their variations and extensions, for diagnosing schizophrenia disorder. However, single modality like sMRI is employed in this paper.

### 4.3 Deep Learning

DL has rapidly become the dominant paradigm for medical imaging analysis. This section includes the contribution of previous papers for SCZ diagnosis using DL models as shown in Table 3.

Li et al. [47] segmented sMRI into GM, WM, and CSF and converted GMV into 2D slices. These slices are fed to CNN for classification. The authors are constrained only to GMV for the SCZ diagnosis and ignore the WMV and CSF parts of the brain. Bagherzadeh et al. [48] and paper [52] calculated the connectivity matrix using EEG signals of SCZ and HC patients, and converted it to a 2D image. Image is fed to a hybrid CNN-(Long short-term memory) LSTM DL network for classification. Although authors [48] achieved a very good ACC of 99.93%, experiments have been done on a small dataset. Also, high computational power is required to train CNN and LSTM DL networks. In another paper, Wu et al. [49] used recurrent autoencoder (RAE) and a fully connected classifier to capture the EEG patterns of SCZ and HC subjects. The proposed method is simple to implement but the size of the dataset used in this study is very small and the authors did not investigate the generalization capability of RAE. [50] investigated the EEG signals by the concatenation of extracted features using machine intelligence and using deep learning. The concatenated features are classified using LG, SVM, and DT classifiers. The proposed approach converts EEG signal into scalograms and spectrograms, increasing the model's computational complexity. Authors in [51] converted 1D EEG signals into 2D scalogram images using continuous wavelet transform to explore temporal-frequency features of the EEG signal. The resulting scalogram images are fed to the VggNet16-based CNN network for feature extraction and classification. In the proposed model, as scalograms are the CNN model's input, the various hyperparameters should be appropriately selected for the optimum performance results.

Previous papers [53, 54, 68] have employed a neural network-based methodology to identify structural and dynamic functional relationships. Paper [54] have discussed functional connectivity of brain whereas [68] have used sMRI modality and [53] have discussed multimodality approach. Moreover, [68] shows the best performance of 84.4% is achieved. However, paper [53] shows limitation with the resting state of brain pattern and affect the overall diagnosis performance. Hu et al. [61, 66] and [62, 72] proposed a 3$D$ CNN modal for classification of sMRI images of SCZ. Paper [62] illustrates the best ACC of 97%. However, work showed demerits like overfitting problems and others. Similarly, papers [57, 69] have taken multimodal and fMRI approaches, respectively, and [69] has shown the best ACC of 99.35%.

**Table 2** Machine Learning Methods

| Input type | Reference | Dataset | Modalities | Feature Extracted | Methods | Performance(%) |
|---|---|---|---|---|---|---|
| 2D scans | Filipovych et al. [17], 2012 | Clinical/SCZ-71, HC-72 | sMRI | GM | JMMCC | CI-95 |
| | Ulas et al. [18], 2012 | Clinical/SCZ-50, HC-50 | sMRI | AD,TL and Th | MKL-SVM | ACC-81 |
| | ArivuSelvan and Moorthy [19],2020 | / SCZ-115, HC-76 | sMRI | Th | ANN | MSE-0.83 |
| | Park et al.[20],2020 | Clinical/SCZ-86,HC-66 | sMRI | HP | LR, AB, XGBoost, and SVM | ACC-80.4 |
| | Febles et al.[21],2022 | Clinical/SCZ-54,HC-54 | EEG | - | Multi-kernel SVM | ACC-83 |
| | Rosa et al.[22],2022 | Clinical/SCZ-20,HC-20 | Post-mortem brain | HP and DLPFC | RF | AUC-95 |
| 3D scans | Ulas et al. [23],2011 | Clinical/SCZ-59,HC-55 | sMRI | AD and Cortex region | Linear SVM | ACC-85.96, |
| | Castellani et al. [24], 2012 | Clinical/SCZ-54,HC-54 | MRI | DLPFC | SVM | ACC-84.09 |
| | Modinos et al. [25], 2012 | Clinical/SCZ-40 | fMRI | Insula, FL and AD | Linear SVM | ACC-75 |
| | Arbabshirani et al. [26],2013 | Clinical/SCZ-28, HC-28 | fMRI | Whole brain | KNN,SVM,DT and ANN | ACC-96 |
| | Castro et al. [27], 2014 | Clinical/SCZ-31 HC-21 | fMRI | Whole brain | SVM and MKL | ACC-90 |
| | Jafri and Calhoun [28],2016 | Clinical/SCZ-20,HC-15 | fMRI | FL, PL, OL and TL | Feed forward Neural network | ACC-76 |
| | Bae et al. [29],2018 | OpenNeuro/SCZ-21, HC-54 | fMRI | Whole brain | SVM, DT, KNN and LDA | ACC-92.1 |
| | Chin et al. [30],2018 | Clinical/SCZ-141,HC-71 | sMRI | GM | Regularized SVM | ACC-92 |
| | Nimkar et al. [31], 2018 | COBRE/ - | fMRI | Whole brain | SVM, KNN, RF and LDA | ACC-94.12 |
| | Pierrefeu et al. [32], 2018 | Clinical/ SCZ-37 | fMRI | FL and TL | SVM | AUC-73 |
| | Li et al.[33],2019 | COBRE/SCZ-60,HC-71 | fMRI | Whole brain | KNN, LDA and SVM | ACC-76.34 |
| | Liu et al.[46],2019 | Clinical/SCZ-28, HC-28 | fMRI | Whole brain | SVM | ACC-92.9 |
| | Sutcubasi et al.[36],2019 | Clinical/SCZ-39,HC-23 | DTI | Whole brain | ANN | ACC-81.25 |
| | Vyškovský et al. [37], 2019 | Clinical/SCZ-52, HC-52 | sMRI | GM | MLP | ACC-73.12 |
| | Chen et al.[38],2020 | COBRE/SCZ-34,HC-34 | sMRI | GM and WM | Linear SVM | ACC-85 |
| | Lei et al.[39],2020 | COBRE/SCZ-295,HC-452 | sMRI and fMRI | GM and WM | SVM | ACC-90.83 |
| | Vieira et al. [40],2020 | Clinical/SCZ-514,HC-444 | sMRI | GMV, Cortical thickness | SVM, KNN, LR, and DNN | ACC-70 |
| | Yang et al.[41],2020 | OpenfMRI/SCZ-44,HC-44 | fMRI | Whole brain | SVM | ACC-99.46 |
| | Yassin et al.[42],2020 | Clinical/ SCZ-97 | sMRI | Cortical thickness, surface area, and subcortical volume | SVM, RF, LG, AB, DT and KNN | ACC-75 |
| | Lin et al.[43],2021 | Clinical/SCZ-141,HC-69 | fMRI and DTI | Whole brain | Multi-kernel SVM | ACC-95.33 |
| | Shi et al. [34],2021 | COBRE/SCZ-71,HC-74 | sMRI and fMRI | GM | LDA | ACC-93.75 |
| | Algumaei et al.[44],2022 | COBRE/SCZ-70,HC-70 | fMRI | HP, TL and FL | SVM | ACC-98.57 |
| | Tanveer et al.[45],2022 | COBRE/SCZ-72,HC-74 | sMRI | GM and WM | SVM and RF | ACC-80.71 |

**Table 3** Deep Learning Methods

| Input type | Reference | Dataset | Modalities | Methods | Performance(%) |
|---|---|---|---|---|---|
| 2D scans | Li et al.[47], 2021 | Clinical / SCZ-89, HC-83 | sMRI | CNN | ACC-99.72 |
| | Bagherzadeh et al. [48], 2022 | RepoD / SCZ-14, HC-14 | EEG | CNN-LSTM | ACC-99.93 |
| | Wu et al. [49], 2022 | Clinical / SCZ-14,HC-14 | EEG | RAE | ACC-81.81 |
| | Nsugbe et al.[50], 2022 | SCZ-10, HC- 10 | EEG | CNN | ACC-98 |
| | Aslan and Akin[51], 2022 | Mental health rehabilitation centers (MHRC) and Clinical / SCZ-14,HC-14 | EEG | CNN | ACC-99.5 |
| | Sharma et al. [52], 2022 | Clinical / SCZ-45,HC-39 | EEG | CNN and LSTM | ACC-96 |
| 3D scans | Calhoun et al. [53], 2017 | Clinical / SCZ-144,HC-154 | sMRI and fMRI | Feedforward neural network | Alignment score-4.3 |
| | Han et al. [54], 2017 | Clinical / SCZ-39,HC-31 | fMRI | Feedforward BP neural networks | ACC-79.3 |
| | Plis et al. [55], 2018 | Clinical / SCZ-144,HC-154 | sMRI and fMRI | DNN | - |
| | Zeng et al. [56], 2018 | Clinical / SCZ-474,HC-607 | fMRI | DANS | ACC- 85 |
| | Oh et al. [57], 2019 | Clinical / SCZ-103,HC-41 | fMRI | 3D CNN | ACC- 84.43 |
| | Pominova et al.[58], 2019 | OpenNeuro / SCZ-50,HC-122 | fMRI | CNN | Mean- 82.3 |
| | Srinivasagopalan et al. [59], 2019 | Clinical / SCZ-69,HC-75 | sMRI and fMRI | DNN | ACC-94.44 |
| | Yan et al. [60], 2019 | Clinical / SCZ-558,HC-542 | fMRI | RNN | ACC-83.2 |
| | Hu et al. [61], 2020 | NUSDAST and IMH/SCZ-289,HC-210 | sMRI | 3D CNN | ACC- 79.27 |
| | Oh et al. [62], 2020 | Schiz Connect / SCZ-424,HC-449 | sMRI | 3D CNN | ACC-97 |
| | Smucny et al. [63], 2020 | Clinical / SCZ-139,HC-138 | fMRI | MLP | ACC- 70 |
| | Wang et al. [64], 2020 | COBRE / SCZ-60,HC-71 | fMRI | MKCN | ACC-82.4 |
| | Gagana et al. [65], 2021 | COBRE / SCZ-69,HC-75 | sMRI and fMRI | Gaussian process classifier and H2O | AUC- 91 |
| | Hu et al.[66], 2021 | NUSDAST and IMH / SCZ-289,HC-210 | sMRI | 2D and 3D CNN | ACC- 79.27 |
| | Kandry et al.[67], 2021 | OpenNeuro / Subjects-99 | fMRI | VGG16 | Mean- 94.33 |
| | Korda et al. [68], 2021 | Clinical / SCZ-141,HC-238 | sMRI | Neural network | ACC-84.4 |
| | Masoudi et al. [69], 2021 | COBRE / SCZ-64,HC-81 | sMRI, fMRI and DTI | 3D CNN | ACC-99.35 |
| | Zhao et al. [70], 2021 | Clinical / SCZ- 558,HC-542 | fMRI | Convolutional RNN | ACC-85 |
| | Cui et al. [71], 2022 | Clinical / SCZ-662,HC-613 | sMRI | DNN | ACC-85.74 |
| | Wen et al. [72], 2022 | Clinical / SCZ-38,HC-39 | sMRI | 3D CNN | ACC-74.9 |
| | Patro et al. [73], 2022 | Online / SCZ-300,HC-300 | sMRI | Light weighht 3D CNN | ACC-92.22 |
| | Zhang et al. [74], 2022 | Clinical / SCZ-437,HC-450 | sMRI | 3D CNN | ACC-92 |

Cui et al. [71] have applied deep neural networks (DNN) to classify the SCZ and HC. Performance using the SVM classifier consists of an ACC of 85.74%. Similarly, Amin et al. [55], and paper [59] have taken multimodal images and showed algorithm higher ACC of 94.44%. Wang et al. [64] have discussed a Multi-kernel capsule network (MKCN) to analyze the disease and achieved an ACC of 82%. Gagana [65] has provided insights regarding multiple automated DL techniques with an AUC of 0.88. Zeng et al. [56] have collected datasets from multi-site and employed discriminant autoencoder (DANs) network with sparsity constraints SCZ detection. Smucny et al.[63] have assessed six ML techniques to predict the response rate based on the fMRI feature employing the brief psychiatric rating scale (BPRS) whereas [67] used Visual Geometry Group (VGG) network. Yan et al. [60] and paper [70] have proposed recurrent neural network (RNN) model. Paper [60] categorizes 558 SCZ and 542 HC individual with 83.2% ACC. Moreover, it provides advantages because of spatio-temporal information. Marina Pominova et al. [58] have explored the potential of 3$D$ deformable convolution, which is implemented on residual volumetric network architecture. However, it showed limitations in the computational cost term of deformable convolution. Patro et al.[73] extracted spatial and spectral data using CNN, and an ensemble bagging classifier is employed for classification. In contrast to previous classification techniques, 3D CNN exhibits effective performance and can distinguish between control and SCZ data. However, training 3D models costs more money computationally, but they can extract more information from high-dimensional voxels than 2D models. Zhang et al. [74] have highlighted the potential of DL to enhance schizophrenia diagnosis and recognise the structural neuroimaging characteristics of the disorder from T1-weighted brain data. Additionally, the GPU memory demand significantly restricts the complexity of the models.

### 4.4 Discussion

SCZ is a psychiatric disorder that causes cognitive impairment and affects the brain's general functioning, characterized by positive and negative symptoms. Structural and functional brain anomalies in schizophrenia have recently been confirmed by MRI [20, 29, 61]. The neuroscience of this condition is now receiving a lot more attention due to these discoveries, which have broadened the scope of both clinical and fundamental scientific studies. Schizophrenia MRI findings include (1) ventricular elongation; (2) involvement of medial temporal lobe structures (amygdala, hippocampus, parahippocampal gyrus); (3) involvement of STG; (4) involvement of parietal lobe structures (particularly the inferior parietal lobule and its subdivision into angular and supramarginal gyrus); and (5) involvement of subcortical brain regions. The prevalence and distribution of anomalies point to a disruption in the normal development of neural connections within and across brain areas. Failure to activate a particular area under certain circumstances suggests insufficient circuitry. The focus of SCZ research has shifted from cognition to emotion, social cognition, motivation, and the reward system.

Abnormal frontotemporal activity is linked to downstream processes. Verbal learning problems are common in SCZ, and fMRI studies have shown anomalies in frontotemporal circuits during learning. In SCZ, the frontal brain is underactive, particularly the inferior prefrontal area [75]. Temporal lobe as ab ROI based assessments is inconsistent. Most investigations have shown reduced activity in the hippocampus and parahippocampal gyrus. Performance may be a factor since studies have different approaches to it. Working memory and cognitive control are impaired in SCZ, according to fMRI. Hippocampal impairment commonly accompanies prefrontal dysfunction, indicating SCZ disrupts the frontotemporal connection. Advances in image analysis allow evaluation of distributed brain circuitry and testing of network model ideas.

In this paper, ML approaches like DT, SVM, and ANN algorithms are introduced for SCZ diagnosis. ANN networks can learn according to the training data and apply their experience in presenting the discrete value result and providing immunity from slight noise or error. It is also suitable for multi-class data. ANN approach, however, shows overfitting problems. Moreover, methodology results in traps in local minima and cannot provide a global outcome [76]. The black-box view of ANN does not give a clue about network functionality, so researchers lost interest in ANN. Another traditional ML algorithm, SVM, effectively separates the semi-structured and disorganized data and manages the high-dimensional data. Also, this algorithm is less complicated and operates well with small datasets. SVM technique depends on the kernel, and the final output varies according to the kernel function. When we compare the SVM with ANN, the risk of overfitting is less in SVM, and it shows better ACC compared to the ANN.

Moreover, ANN [19, 37] dependency on a large dataset for optimum result reduce its demand as a classifier where state-of-the-art ML technique like SVM [35, 43] showed the better performance with the large dataset and its computational simplicity rise the applicability. DL [66, 68] networks have attracted most researchers for accurate, automatic, and fast classification of the SCZ disease. On comparing traditional state-of-the-art ML and trending DL models for classification of SCZ and HC brain images. However, each has its own merits and demerits. Manual feature extraction leads to an in-depth analysis of the specific regions taken into consideration for ML algorithms which is not possible for DL approaches, whereas, DL algorithms extract all level of features from the given input without human intervention.

Many researchers have widely adopted automated classification opting for ML and DL approaches as they provide a trustworthy opinion about the disease. In handling the large dataset, DL is the best and yields the highest ACC compared to traditional ML methods. The findings of this review lead us to believe that DL models demonstrate increased ACC with a trade-off with computational power. The new integrated methodology of ML techniques will play an increasingly important role in the future, both for early diagnosis and a more accurate examination of treatment and for determining the long-term prognosis of patients with SCZ. To recap, neuroimaging research in psychiatry employing deep learning is still growing to improve performance. While significant obstacles are ahead, our observations provide preliminary evidence for both ML/DL possible role in developing biological neuroimaging biomarkers for mental illnesses.

# 5 Challenges and Future scope

In this section merits and limitations of ML techniques, along with the future scope, is discussed:

1. *Modality*: As SCZ includes structural as well as functional atrophies in brain regions, no single modal data can provide all information about the disease. Hence, multimodal approaches play a crucial role in understanding the functional, metabolically, structural and morphological information. These studies only include the findings on MRI, DTI, EEG, and fMRI images to diagnose SCZ. Further work can be extended to include other imaging modalities like positron emission tomography (PET) and clinical studies. Including a multimodal approach will provide robustness to the learning model; however, computational costs will be high. Another approach to handle computational power is by implementing fused images as input. The fusion of multimodal data like MRI and fMRI to see the structural and functional changes in the brain, thereby providing complementary data to the model, which can indirectly help in better understanding the disease.

2. *Schizophrenia detection*: The drawback with SCZ is regarding the follow-up analysis and behavior prediction. A more comprehensive cortical parcellation may be necessary to link the areas that committed the most to DL with regions identified in prior research using voxel-based techniques. Further, the networks can also be used to segment different brain regions from 3D MRI images. By segmenting the brain regions, we can observe the changes in these regions with the progression of the disease.

3. *Challenges in machine learning with future scope*: For accurate SCZ prediction, selective feature extraction and selection model is required which can result in efficient SCZ biomarkers. ML implementation necessitates a level of expertise in the AI field such as choosing the optimal algorithms for each segment and is challenging to obtain an accurate diagnosis of SCZ. For preprocessing of sMRI and fMRI modalities, it is yet another challenge to obtain the best learnable dataset for the selected ML algorithms.

4. *Challenges in deep learning with future scope*: To improve the efficacy of the disease diagnosis, various Ensemble DL-based networks can be implemented. Ensemble-based learning provides diversity to the learning model, directly improving the model's performance. Also, in DL networks, hyperparameters are essential in training the network. Therefore, further work can extend to optimize the hyperparameters of DL networks using metaheuristic algorithms. Another extension of DL is including a shallow network-based classifier approach over a probabilistic-based classifier.

# 6 Conclusion

This review provides evidence of changes in brain patterns in schizophrenia using structural MRI, functional MRI (fMRI), diffusion tensor imaging, and electroencephalogram (EEG). A practical and precise diagnosis of psychiatric disorders is crucial for initiating and selecting effective treatment. The paper ensures that machine learning/deep learning on neuroimaging is a promising tool for developing diagnostic models that could benefit the diagnosis of SCZ. Moreover, the application of deep learning in neuroimaging for psychiatric disorders has shown favorable results and obtained better performance than conventional state-of-the-art machine learning (ML) techniques. This review provides the research gap and future scope of the work which help researchers to develop new models for their upcoming contribution to a schizophrenia diagnosis.